\documentclass[10pt,letterpaper, table]{article}

\newcommand{\viipt}{\fontsize{7pt}}

\newcommand{\ixpt}{\fontsize{9pt}}
\newcommand{\xipt}{\fontsize{10.95pt}}
\newcommand{\xiipt}{\fontsize{12pt}}
\newcommand{\xivpt}{\fontsize{14.4pt}}
\newcommand{\xviipt}{\fontsize{17.28pt}}
 
\usepackage{bm}
\usepackage{enumerate}
\usepackage{amsmath,amsfonts,amsthm}
\usepackage{algorithm, algorithmic}
\usepackage{booktabs}
\usepackage{multirow}
\usepackage{setspace}
\usepackage{fancyhdr}
\usepackage{lastpage}
\usepackage{extramarks}
\usepackage{chngpage}
\usepackage{soul,color}
\usepackage{graphicx,float,wrapfig}
\usepackage{xspace}

\usepackage{todonotes}
\usepackage{array}
\usepackage{xcolor}
\usepackage{hhline}
\usepackage{colortbl}

\usepackage{cogsci}
\usepackage{pslatex}
\usepackage{apacite}

\makeatletter
\DeclareRobustCommand\onedot{\futurelet\@let@token\@onedot}
\def\@onedot{\ifx\@let@token.\else.\null\fi\xspace}

\def\eg{\emph{e.g}\onedot} 
\def\ie{\emph{i.e}\onedot}

\def\Exp{Exp\onedot}
\makeatother

\newcommand{\smallvs}{\vspace{-10pt}}

\definecolor{g0}{rgb}{0,0,0}
\definecolor{g1}{rgb}{0.2,0.2,0.2}
\definecolor{g2}{rgb}{0.35,0.35,0.35}
\definecolor{g3}{rgb}{0.5,0.5,0.5}
\definecolor{g4}{rgb}{0.65,0.65,0.65}
\definecolor{g5}{rgb}{0.8,0.8,0.8}

\definecolor{MyRed}{rgb}{1.0,0.0,0.0}
\definecolor{MyBlue}{rgb}{0,0.08,1}

\newcolumntype{L}[1]{>{\raggedright\let\newline\\\arraybackslash\hspace{0pt}}m{#1}}
\newcolumntype{C}[1]{>{\centering\let\newline\\\arraybackslash\hspace{0pt}}m{#1}}
\newcolumntype{R}[1]{>{\raggedleft\let\newline\\\arraybackslash\hspace{0pt}}m{#1}}

\newlength\Origarrayrulewidth

\newcommand{\Hline}{%
  \noalign{\global\setlength\Origarrayrulewidth{\arrayrulewidth}}%
  \noalign{\global\setlength\arrayrulewidth{3pt}}\arrayrulecolor{red}\hhline{|~|~|~|~|-|~|~|}\arrayrulecolor{black}%
  \noalign{\global\setlength\arrayrulewidth{\Origarrayrulewidth}}%
}

\newcommand\Thickvrule[1]{%
  \multicolumn{1}{!{\color{red}\vrule width 2pt}c!{\color{red}\vrule width 2pt}}{#1}%
}

\newcommand{\specialcell}[2][c]{%
  \begin{tabular}[#1]{@{}c@{}}#2\end{tabular}}

\title{A Comparative Evaluation of Approximate Probabilistic Simulation and Deep Neural Networks as Accounts of Human Physical Scene Understanding}
\author{{\large \bf Renqiao Zhang$^{1*}$ \quad Jiajun Wu$^{1*}$ \quad Chengkai Zhang$^1$ \quad William T. Freeman$^{1,2}$ \quad Joshua B. Tenenbaum$^1$} \\ $^1$Massachusetts Institute of Technology\quad $^2$Google Research \qquad \{andy17, jiajunwu, ckzhang, billf, jbt\}@mit.edu}

\begin{document}
\maketitle
\footnotetext{$*$ indicates equal contributions.}

\begin{abstract}
Humans demonstrate remarkable abilities to predict physical events in complex scenes.  Two classes of models for physical scene understanding have recently been proposed: ``Intuitive Physics Engines'', or IPEs, which posit that people make predictions by running approximate probabilistic simulations in causal mental models similar in nature to video-game physics engines, and memory-based models, which make judgments based on analogies to stored experiences of previously encountered scenes and physical outcomes. Versions of the latter have recently been instantiated in convolutional neural network (CNN) architectures.  Here we report four experiments that, to our knowledge, are the first rigorous comparisons of simulation-based and CNN-based models, where both approaches are concretely instantiated in algorithms that can run on raw image inputs and produce as outputs physical judgments such as whether a stack of blocks will fall.  Both approaches can achieve super-human accuracy levels and can quantitatively predict human judgments to a similar degree, but only the simulation-based models generalize to novel situations in ways that people do, and are qualitatively consistent with systematic perceptual illusions and judgment asymmetries that people show. 

\textbf{Keywords:} 
physical scene understanding; neural network; analysis by synthesis; simulation engine; blocks world
\end{abstract}

\section{Introduction}
The outputs of vision include not only the objects in a scene and their spatial relations, but also their physical properties and relations: What is heavy or light?  What is balanced or attached, and what isn't?  What is likely to fall? What will happen next? When objects move, their motion can be predicted from these physical inferences; motion can also affect our physical judgments when objects move in unexpected ways.  

These capacities for physical scene understanding are basic to how we see the world. Precursors to them can be found in infants as young as 3-5 months old, even before children acquire their first words labeling kinds of objects~\cite{carey2009origin,baillargeon2004infants}. Building computational models of these abilities has been a target for recent work in both cognitive science and computational vision \cite{battaglia2013simulation,blocksworldrevisited,mottaghi2015newtonian,fragkiadaki2015learning,zheng2015scene,li2016fall}. In contrast to earlier work on intuitive physics that emphasized explicit reasoning about textbook-style physics problems~\cite{mccloskey1983intuitive}, with models focused on people's qualitative judgments~\cite{forbus1984qualitative,siegler1976three}, recent studies of physical scene understanding have looked at more rapid, perceptual inferences, which can be parametrically manipulated and modeled quantitatively, and which could serve as the basis for grounded action planning. Several studies have argued that rapid perceptual inferences about the physics of scenes can be explained by positing an ``intuitive physics engine'' (IPE), a mental system for approximate probabilistic simulation analogous to those used in video-game physics engines~\cite{sanborn2013reconciling,gerstenberg2012noisy,smith2013sources}. These simulation engines approximate object dynamics interacting under Newtonian or other forms of classical mechanics over short time scales, in ways that are perceptually reasonable (if not necessarily physically accurate) and efficient enough to run in real time for complex scenes. 

Other authors have suggested that the simulation-based IPE scheme might be prohibitively expensive for brains to implement~\cite{davis2016scope}. An alternative class of models has been proposed based on stored memories of experienced scenes and physical outcomes, together with pattern recognition algorithms (such as neural networks) for accessing appropriate memory items to predict outcomes in a new scene context~\cite{sanborn2013reconciling,sanborn2014testing}.  

Although cognitive scientists have yet to seriously test memory-based alternatives to simulation in physical scene understanding tasks, AI researchers at Facebook recently demonstrated such a possibility in a working system.  \citeA{lerer2016learning} trained convolutional neural networks (CNNs) to make physical predictions directly from images, judging whether a stack of blocks will fall, as~\citeA{battaglia2013simulation} studied empirically and modeled using approximate probabilistic simulation. Their neural network, named PhysNet, was partly pretrained on ImageNet~\cite{krizhevsky2012imagenet} and then trained on a large dataset of synthetic scenes and outcomes. It achieved a high accuracy ($89\%$) on the task, generalized to real images reasonably well ($67\%$), and exhibited positive correlations with human responses. This suggests that memory-based systems for visual intuitive physics may be promising at least in AI applications, and perhaps also as cognitive models. 

Motivated by the success of CNNs in machine vision object recognition tasks ~\cite{krizhevsky2012imagenet}, neuroscientists have proposed analogous architectures as accounts of the fast feedforward aspects of human visual object recognition~\cite{yamins2014performance,serre2007feedforward}.  If CNNs can be successfully applied to physical scene understanding tasks as well, they could offer a compelling alternative to simulation as an account of how people can predict physical outcomes so well, so quickly. 

Our goal in this paper is to conduct the first rigorous empirical comparisons of simulation-based (IPE) and neural-network-based (CNN) models for physical scene understanding. Although CNNs have many appealing features as models of visual cortex, they also have features that are less appealing -- and arguably less human-like.  They typically require large amounts of training data, which a human might not have access to. Large training sets may be required for any new scenario, even if it is just a simple variation on previously seen cases. For instance, in order to predict whether a pile of four blocks is stable, a CNN may have to see at least thousands of cases that either do or do not fall under gravity. In contrast, an IPE model, just like humans, is able to make many predictions with reasonable accuracies without training, as the simulation engine within encodes abstract physical knowledge that applies to a very wide range of scenes.

Even with a large amount of training data, it is unclear whether the knowledge learned by CNNs may be transferable to some similar cases. \citeA{lerer2016learning} showed that a network trained on images of two and four blocks could generalize to images of three blocks to some extent, but there is no clear way for a neural network to answer a different but related question to those it is trained for, \eg, in which direction the blocks would fall, unless explicit labels are provided during training.  One of the main points in favor of IPE models is their ability to explain how people can easily make many different judgments about very different configurations of blocks, without specific training~\cite{battaglia2013simulation}.

Perhaps most interestingly, people are prone to systematic ``physics illusions'' that IPE models naturally capture.  For instance, stacks of blocks often look to people as if they are sure to fall when they are actually carefully balanced. People do not, however, make the opposite error: They do not systematically mistake unstable stacks for stable ones. Probabilistic simulation-based models are similarly tempted to make this asymmetric pattern of errors~\cite{battaglia2013simulation}: Small amounts of uncertainty in the simulation can make a stable configuration appear unstable, but are unlikely to make an unstable one appear stable. It is unclear whether neural-network-based models can capture these perceptual illusions.

In this paper, we report four experiments comparing the behavior of discriminatively trained neural networks and generative simulation-based models with human judgments on blocks-world physics tasks, addressing the questions above. \Exp1 evaluates the performance of the IPE model and performance-optimized neural networks in predicting block stability. \Exp2 explores the role of limiting CNN training data, to see if performance on smaller training sets looks more human-like. \Exp3 evaluates both model classes for asymmetries in the stability illusions described above. \Exp4 tests CNNs and IPE models' ability to generalize to situations slightly different from those the CNN was trained on. 

\section{The Blocks World}

\begin{figure}[t]
\centering
\begin{tabular}{C{0.46\linewidth}C{0.46\linewidth}}
\includegraphics[width=0.485\linewidth]{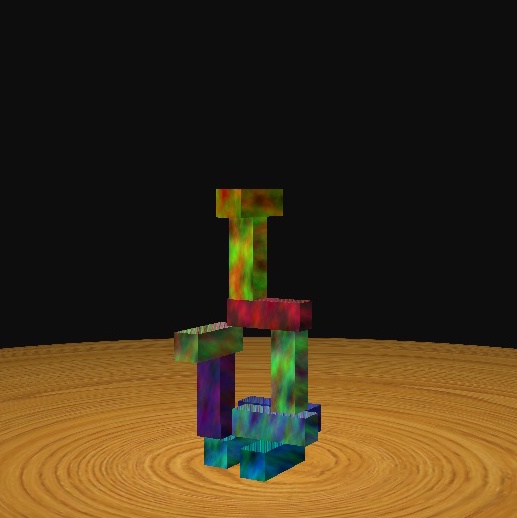}
\includegraphics[width=0.485\linewidth]{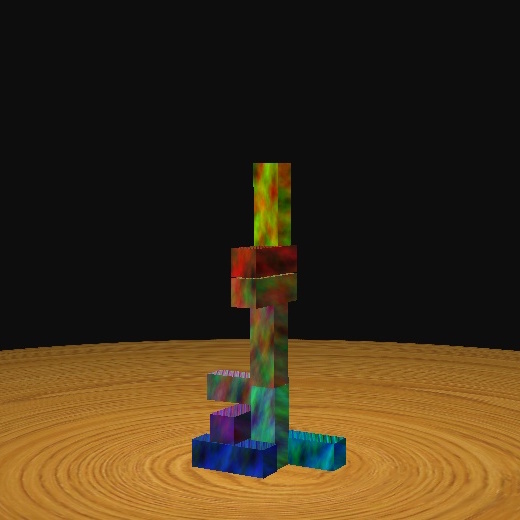} &
\includegraphics[width=0.485\linewidth]{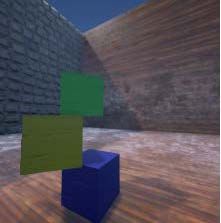}
\includegraphics[width=0.485\linewidth]{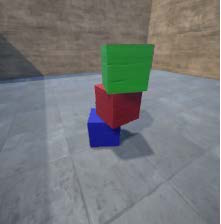} \\
\specialcell{Stimuli in \\\citeA{battaglia2013simulation}} & \specialcell{Stimuli in \\\citeA{lerer2016learning}}
\end{tabular}
\includegraphics[width = 0.96\linewidth]{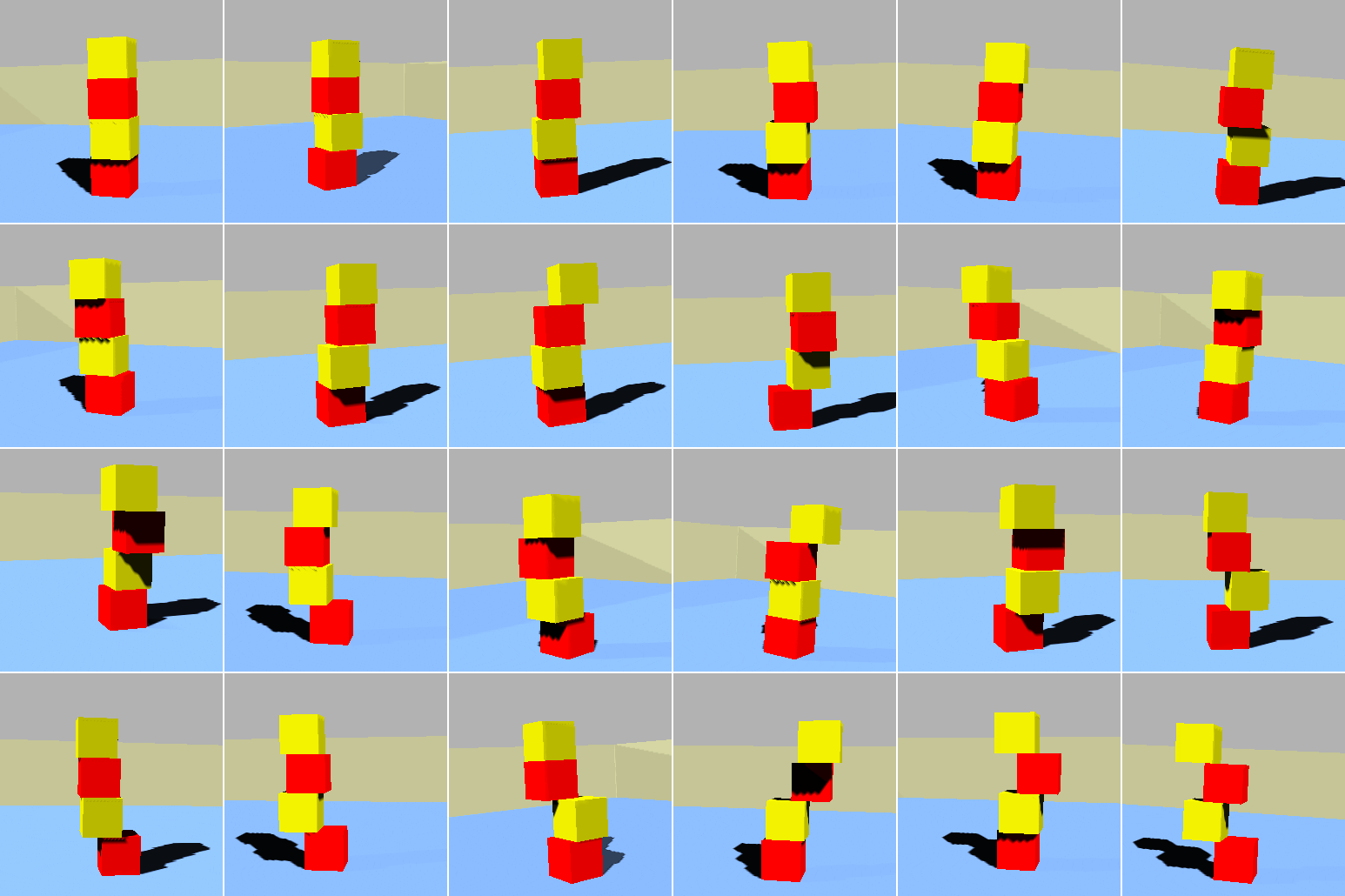}
    \vspace{-10pt}
    \caption{Sample stimuli used by~\citeA{battaglia2013simulation}, Facebook AI Research~\cite{lerer2016learning}, and us. Our stimuli are ordered by increasing visual instability (defined in Experiment 3)}
  \vspace{-15pt}
\label{fig:cubes}
\end{figure}

For our experiments, we study a set of seemingly simple but physically rich scenarios: a pile of blocks with one on top of another. Our goal is to study how humans and computational models behave on various tasks given these stimuli, and to reveal possible correlations between them. We now illustrate our stimuli in detail.

For each stimulus, there are four blocks with side length $1$ meter piled on the ground, each supporting another on top of it. There is only one block at the same height level. Because laying blocks at uniform random is likely ($p=75\%$) to result in an unstable system, we draw the horizontal position of a block from a normal distribution with variance $0.29^2$ centered at the horizontal position of the block under it, to ensure that there are half stable and half unstable piles in the dataset. Later, we study cases where the number of blocks varies, and for them we update the variance accordingly.

Whether blocks are stable, \ie, groundtruth labels, can be derived from the coordinates of blocks. A block will fall if and only if the center of mass of all blocks above it, including itself, does not fall on top of the block under it. 

For rendering, we generate images of resolution $256\times 256$.
We place a pile of blocks in a virtual experiment field with a size of $30\times 30$ meters and a height of $4$ meters. 
We have one light source, 16 meters high, to simulate real-life lightening. We also vary the position, focal point, and tilt angle of the camera. We represent its coordinates in cylindrical coordinates $(r,\theta,z)$, with origin on the ground right beneath the center of the bottommost block. The camera positions are sampled from $r\sim N(11,0.3^2)$, $\theta\sim \text{Uniform}(0,\pi/2)$, and $z\sim N(3,0.01^2)$. We choose these parameters to ensure all blocks are within the view of the camera. The focal point of the camera is set at the center of the pile plus a Gaussian noise with variance $0.2^2$. We also tilt the camera; its angle from the direction of projection is sampled from $N(0,2^2)$. We incorporate these variances for evaluating the generalization ability of the models.

\section{Computational Models}

We study two classes of computational models. One is the Intuitive Physics Engine (IPE) Model~\cite{battaglia2013simulation}, which aims to simulate humans' reasoning on physical scenes by an approximate probabilistic simulation engine. The other is convolutional neural networks (CNNs), a class of discriminative recognition models that have gained much popularity in AI fields like computer vision in recent years.

\begin{figure}[t]
\centering
\includegraphics[width = \linewidth,trim={0 0.3cm 0 0.3cm},clip]{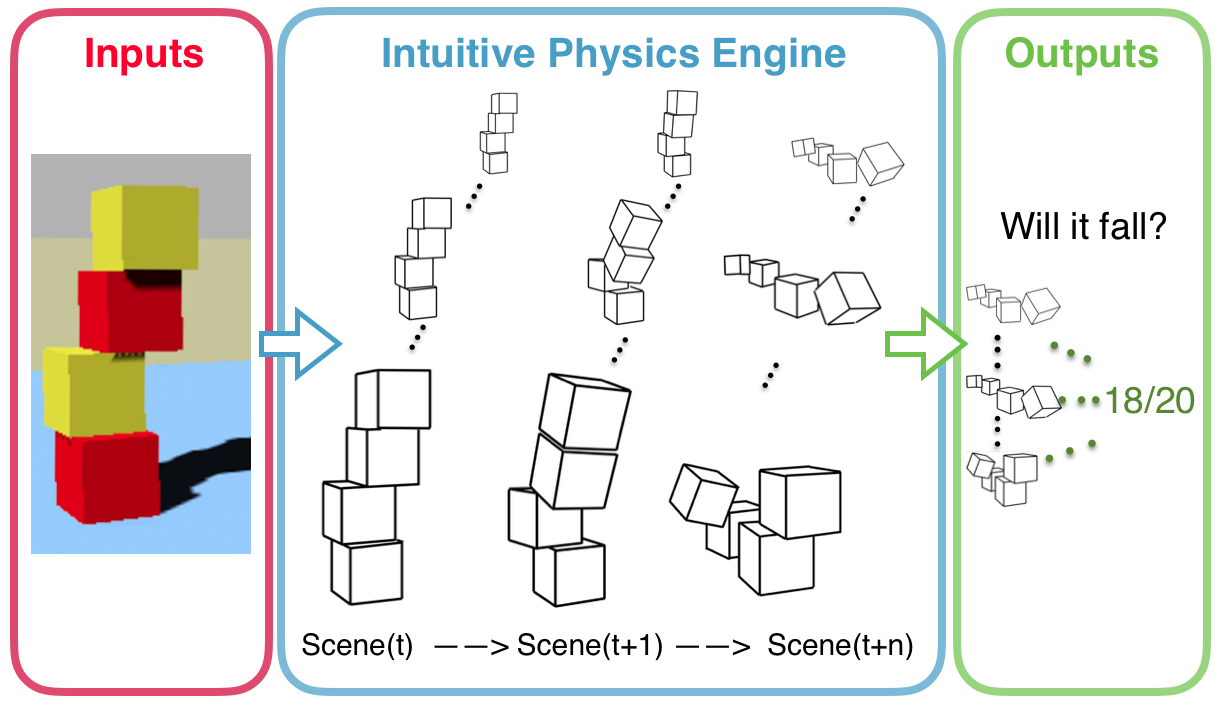}
  \vspace{-20pt}
  \caption{The Intuitive Physics Engine (IPE) model}
\label{fig:ipe}
\vspace{-10pt}
\end{figure}

\subsection{The Intuitive Physics Engine Model} 

The Intuitive Physics Engine (IPE) consists of two components: a Bayesian vision system, which infers the configurations of blocks from given images, and a physical inference system, which calculates the Bayesian posterior probability distribution of physical properties (\ie, stability) by running a number of simulations under perturbation forces and geometric noises. Figure~\ref{fig:ipe} illustrates the IPE model. For more details, please see~\citeA{battaglia2013simulation}.

For each scene, we render images of the initial state under perspective projection from three fixed viewpoints rotated by $45^{\circ}$. These triplets of images are then fed into the Bayesian vision system, which uses a Metropolis-Hasting (MH) sampling algorithm to infer a Bayesian posterior distribution of the scene's initial state (position, height, and the number of blocks presented). We run the MH sampling for $5,000$ steps, with a 2D Gaussian blurring kernel of width 2 on the observed images, as suggested by~\citeA{battaglia2013simulation}. 

With the inferred initial geometry, we run 20 simulations for each scene using the Open Dynamics Engine (ODE)~\cite{ode}. We set the friction coefficient to $0.2$, the bounce coefficient to $0.2$, and the side-length and density of each block to $1m$ and $500 kg/m^3$, respectively. Gravity is set to $9.81m/s^2$ pointing downwards. Before each simulation starts, a horizontal zero mean Gaussian noise $\sigma$ is added to the positions of blocks. Then the simulation runs at a step size of $10ms$ for $2$ seconds. During the first second, a horizontal force with magnitude $\phi$ is exerted at the center of the bottom face of the bottommost block. The direction of the force is uniformly sampled from $(0,2\pi)$ and changes at a frequency of 50Hz. We consider a pile unstable if the vertical coordinate of the top block changes by more than $0.2$ meters when the simulation ends.

\subsection{Convolutional Neural Networks} 

\begin{figure}[t]
\centering
\includegraphics[width = \linewidth]{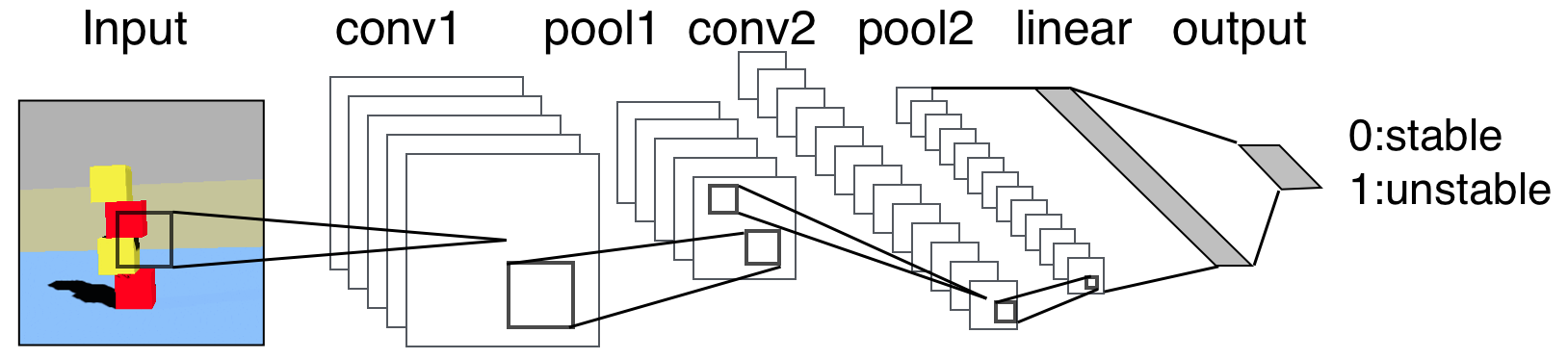}
  \vspace{-20pt}
\caption{The structure of LeNet}
\label{fig:lenet}
\vspace{-15pt}
\end{figure}

CNNs have gained much popularity in computer vision~\cite{krizhevsky2012imagenet}. Here we consider two popular CNN frameworks: the small but powerful LeNet~\cite{lenet}, and the widely used AlexNet~\cite{krizhevsky2012imagenet}.

LeNet, originally proposed for digit recognition, has been widely used as a recognition model in vision because of its effectiveness and simplicity~\cite{lenet}. LeNet consists of two convolutional layers, each followed by a pooling layer and an activation layer. There are then two fully connected linear layers at the end. We modify the final layer so that instead of ten outputs for digit classification, the model now has two output units --- its confidences on whether the blocks will fall or not. Figure~\ref{fig:lenet} shows the structure of LeNet.

The second is the popular AlexNet~\cite{krizhevsky2012imagenet}, which achieves impressive performance on ImageNet classification. 
AlexNet consists of five convolutional, pooling, and activation layers, and three linear layers at the end. 
We evaluate both AlexNet pretrained on ImageNet, as well as AlexNet trained from scratch.

We use Torch~\cite{torch} for implementation. 
We set the learning rate to $0.01$ for LeNet and for fine-tuning AlexNet, and to $0.2$ for training AlexNet from scratch. We use stochastic gradient descent for training. 

\section{Behavioral Experiments}

To collect human responses, we first randomly divide all test images into groups, each consisting of $10$ images. We then add four \emph{easy} cases (two stable, two unstable), whose stability is visually apparent, into the group. For each group, we collect $80$ responses on Amazon Mechanical Turk. We only allow workers with an approval rate $>90\%$ to submit responses, and we only accept responses from workers that answered all four easy cases correctly.

\section{Experiment 1: Predicting Falling Blocks}

In our first experiment, we test the performance of the IPE model and neural networks on images with four blocks, and compare the results with human responses.

\smallvs
\paragraph{Experimental Setup}

For the IPE model, we consider cases with various levels of geometric Gaussian noises $\sigma$ and external forces $\phi$ during  physical simulations. We then compare their performance with LeNet, AlexNet, and humans.

We use $1,000$ test images, each with a pile of four blocks. For neural networks, we build a training set of $200,000$ images (disjoint from the test set) with groundtruth labels.

\smallvs\paragraph{Results and Discussions}

\begin{table}[t]
\centering
\begin{tabular}{|l|l|c|c|c|c|c|}
\hline
\multicolumn{2}{|c|}{} & \multicolumn{5}{c|}{$\phi$}\\ 
\cline{3-7}
\multicolumn{2}{|c|}{} &0 & 35 & 40 & 45 & 50\\
\hline
\multirow{5}{*}{$\sigma$}

&0 & \cellcolor{g1} \color{white} 94.2 & \cellcolor{g1} \color{white} 87.2 
& \cellcolor{g2}  \color{white} 79.5 & \cellcolor{g3} 71.3
& \cellcolor{g4} 63.8 \\
&0.05 &  \cellcolor{g2} \color{white} 91.3 &  \cellcolor{g3} \color{white} 83.4 
&  \cellcolor{g3} \color{white} 76.1 &  \cellcolor{g4} 69.1
&  \cellcolor{g4} 61.8 \\
\Hline
&0.1 &  \cellcolor{g4} 83.2 &  \cellcolor{g4} 75.7
& \Thickvrule{ \cellcolor{g5} 70.3} &  \cellcolor{g5} 62.6
& \cellcolor{g5} 56.4 \\
\Hline
& 0.15 & \cellcolor{g5}  72.2 
& \cellcolor{g5}  66.8 &  \cellcolor{g5} 59.4 
&  \cellcolor{g5} 54.2 &  \cellcolor{g4} 51.2 \\
&0.2 &  \cellcolor{g3} \color{white}  58.5 &  \cellcolor{g3} \color{white} 53.8
&  \cellcolor{g4} 52.1 &  \cellcolor{g3} \color{white} 51.0 
&  \cellcolor{g3} \color{white} 50.9 \\
\hline
\end{tabular}
\ \\ \ \\
\begin{tabular}{|l|c|c|c|c|c|}
\hline
Corr & \cellcolor{g1} \color{white} $\geq0.45$
& \cellcolor{g2} \color{white} $\geq0.54$ & \cellcolor{g3} \color{white} $\geq0.56$
& \cellcolor{g4} $\geq0.58$ & \cellcolor{g5} $\geq0.60$ \\
\hline
\end{tabular}
\caption{Accuracies (\%) of the IPE model with different $\sigma$ and $\phi$, and their correlations with human responses. We use $(\sigma, \phi)=(0.1, 40)$ for following experiments.}
\vspace{-10pt}
\label{tbl:IPE}
\end{table}

As shown in Table~\ref{tbl:IPE}, when no geometric error or external force is added to the IPE model ($\sigma=0,\phi=0$), its results almost always match ground-truths ($94.2\%$ accuracy).  Accuracy decreases as noises increase; however, as previously described in~\citeA{battaglia2013simulation}, we also observe that correlation between IPE responses and human predictions goes up. For the following experiments, we use an IPE model with $(\phi,\sigma)=(0.1,40)$ as it matches human performance in terms of both accuracy and correlation.

We compare results for stable and unstable cases separately, and list them in Table~\ref{tbl:asym}. We observe that human predictions and the IPE model responses have an asymmetric pattern: they perform well on unstable cases, but for images with a stable pile of blocks, their accuracies are much worse. On the contrary, neural networks do not exhibit a similar pattern; they have roughly the same accuracies for both cases. We will revisit this asymmetry more in Experiment 3.

\section{Experiment 2: Limited Training Data}

In our second experiment, we inspect the behaviors of neural networks with different sizes of training sets. As our the IPE model requires only one or a few examples for simulation, its performance does not change with the availability of training data. The same applies to humans.

\begin{table}[t]
\centering
\begin{tabular}{lccc}
\toprule
Method & Stable & Unstable & All \\
\midrule
Human & 38.0 & 92.9 & 65.5 \\
IPE & 40.7 & 99.0 & 70.3 \\ 
\midrule
LeNet $(200$K$)$ & 91.3 &89.0 &90.1 \\
AlexNet $(200$K$)$ & 91.5 & 92.3 & 91.9 \\
AlexNet $($Pretrained, $200$K$)$ & 94.5 & 94.7 & 94.6 \\
\midrule
LeNet $(1,000)$ & 68.0 & 69.3 & 68.7 \\
AlexNet $(1,000)$ & 71.8 & 70.1 & 70.9 \\
AlexNet $($Pretrained, $1,000)$ & 72.5 & 74.2 & 73.4 \\
\bottomrule
\end{tabular}
\caption{Accuracies (\%) of humans, IPE, LeNet, and AlexNet (pretrained and not pretrained), on $200$K or $1,000$ images. The results on $1,000$ images are averaged over five models trained on independently sampled sets.} 
\vspace{-10pt}
\label{tbl:asym}
\end{table}

\smallvs
\paragraph{Experimental Setup}

Instead of using training sets of $200,000$ images, we now only provide the networks with training sets of $100$ to $20,000$ images. For each scale, we sample five training sets independently, train one network on each set, and compute the average of their performance. The other setup is same as that in Experiment 1. 

\begin{figure}[t]
\centering
\includegraphics[width=\linewidth,trim={1cm 0 1cm 0},clip]{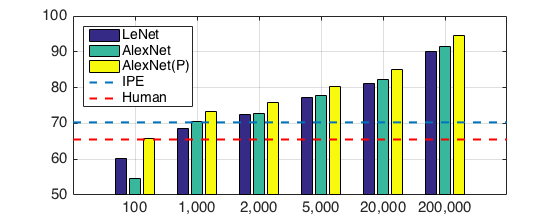}
\vspace{-20pt}
\caption{CNN models with different sizes of training sets}
\vspace{-15pt}
\label{fig:data}
\end{figure}

\begin{figure*}[t]
\centering
\begin{tabular}{ccc}
\includegraphics[width=0.31\linewidth,trim={1.2cm 1.5cm 1.2cm 1cm},clip]{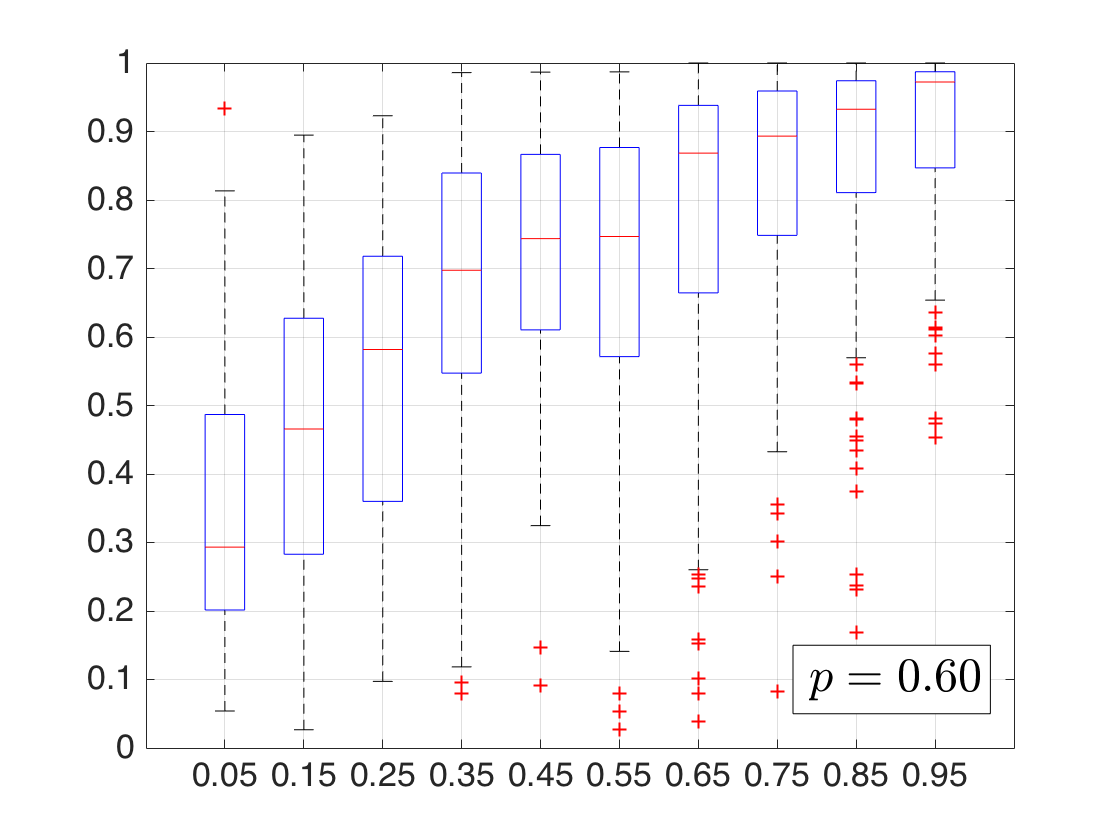} &
\includegraphics[width=0.31\linewidth,trim={1.2cm 1.5cm 1.2cm 1cm},clip]{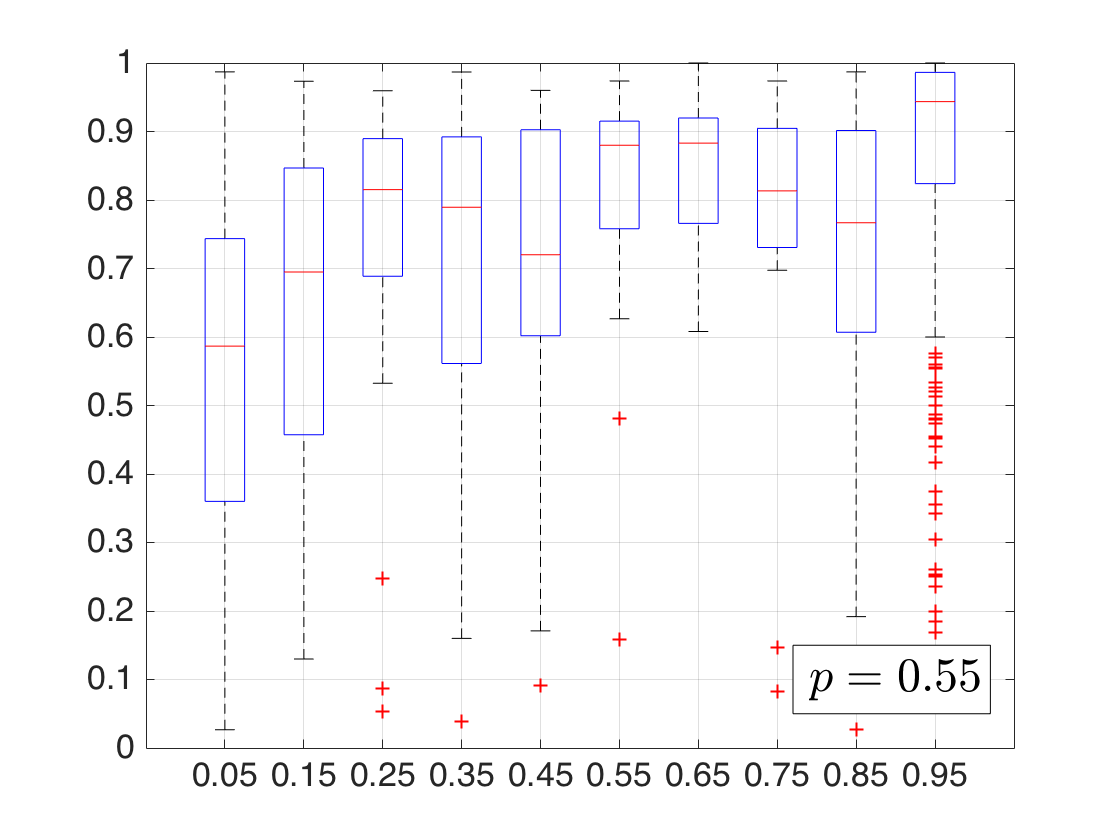} &
\includegraphics[width=0.31\linewidth,trim={1.2cm 1.5cm 1.2cm 1cm},clip]{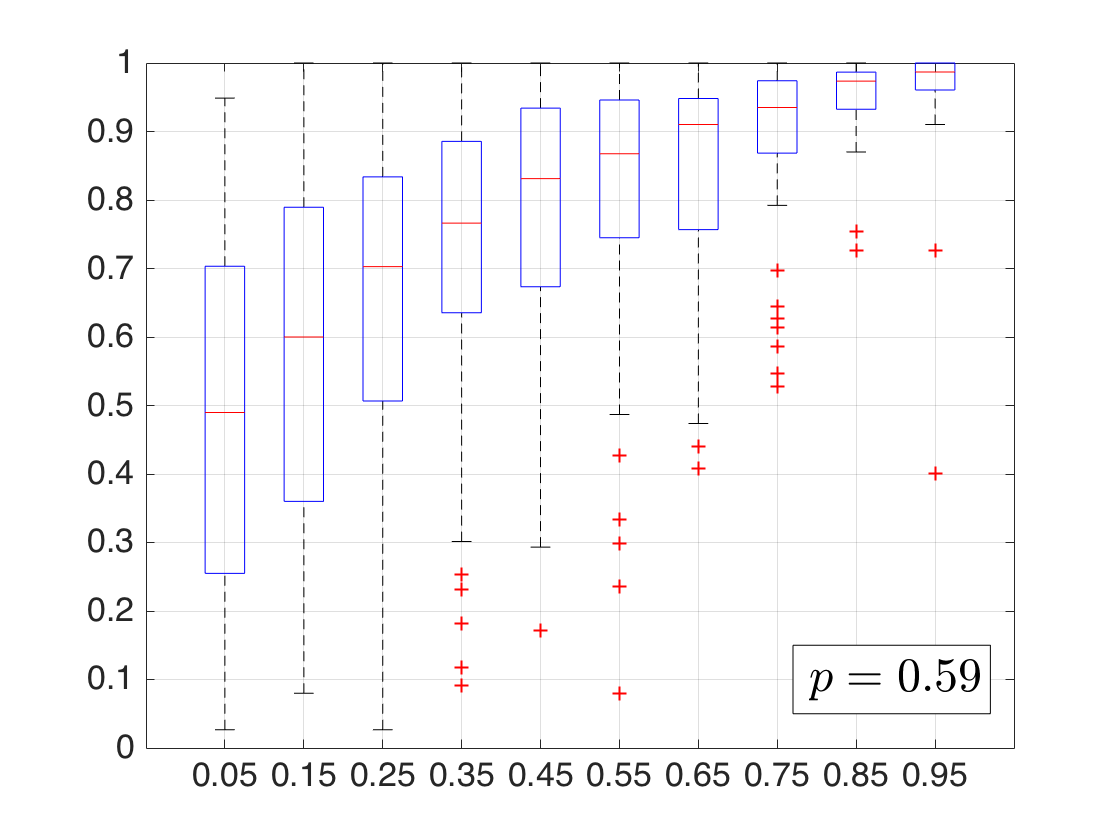}\\
(a) Human ($Y$-axis) vs IPE (X-axis) &
(b) Human (Y) vs LeNet $200$K (X) &
(c) Human (Y) vs LeNet $1,000$ (X) 
\end{tabular}
\vspace{-10pt}
\caption{From left to right: human responses vs (a) responses of IPE (normalized numbers of moving blocks), (b) LeNet trained on the full training set ($200,000$ images), and (c) LeNet trained on $1,000$ images. Results for AlexNet are similar. We list Pearson's correlation coefficients at the bottom-right corner.}
\vspace{-15pt}
\label{fig:IPEvsH}
\end{figure*}

\smallvs
\paragraph{Results and Discussions}

As shown in Figure~\ref{fig:data}, the performance of CNNs decreases as there are fewer training data. 
Although AlexNet (not pretrained) performs better with $200,000$ training images, it also suffers more from the lack of data, while pretrained AlexNet is able to learn better from a small amount of training images.
For our task, both models require around $1,000$ images for their performance to be comparable to the IPE model and humans. 
 
We then evaluate the networks trained with $1,000$ images. As shown in Table~\ref{tbl:asym}, there is still no asymmetric pattern in the responses of the less-trained networks.

We now look into how each model correlates with human responses in more detail. Figure~\ref{fig:IPEvsH} (a) and (b) demonstrate that the IPE model has a stronger correlation with humans, compared to LeNet trained on the full training set. Another interesting finding is that the less-trained LeNet (c) is more human-like. We will discuss this more in the final section.

\section{Experiment 3: Boundary Cases}

We now systematically study the asymmetry we observed in Experiment 1. In particular, we focus on a few  groups with \emph{visually unstable} piles, \ie, piles that are carefully balanced and therefore stable, but illusory to humans so that they believe these blocks will fall. 

\smallvs
\paragraph{Experimental Setup}

\begin{figure}[t]
\centering
\quad
\includegraphics[width=0.215\linewidth]{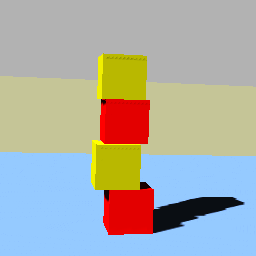}
\includegraphics[width=0.215\linewidth]{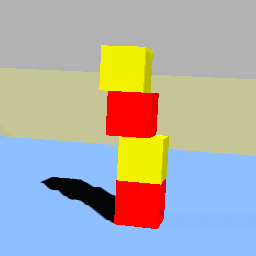}
\includegraphics[width=0.215\linewidth]{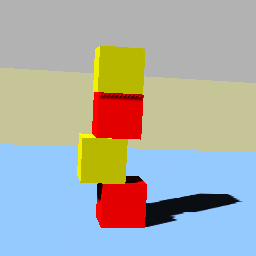}
\includegraphics[width=0.215\linewidth]{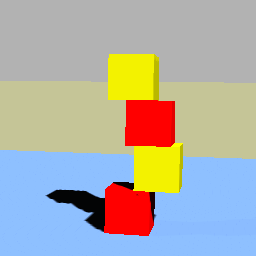}\\
\includegraphics[width=\linewidth,trim={2cm 0 2cm 0},clip]{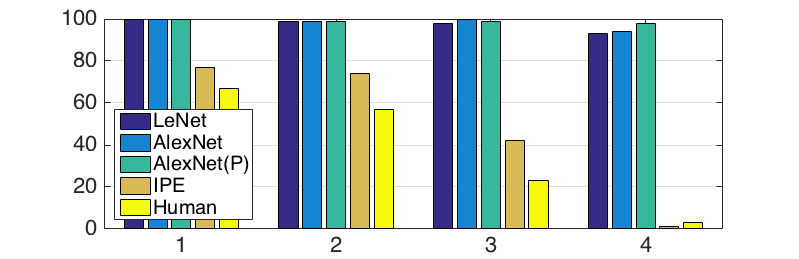}
\vspace{-25pt}
\caption{Upper: stable blocks with increasing visual instabilities; Lower: performance of LeNet, AlexNet, pretrained AlexNet, IPE, and humans on the four datasets. Neural networks are trained on $200$K images. Behaviors of networks trained on a smaller set ($1,000$ images) are similar.}
\vspace{-20pt}
\label{fig:boundary}
\end{figure}

We define \emph{visual instability}, scaling from 0 to 5, to describe how unstable a pile of blocks looks like. A pile with instability value $x$ means there exists at least one block so that the center of mass of the blocks above it lies $x/10$ meters away from its center on x-y plane. 
As the side-length of blocks is 1 meter, a pile with a visual instability value 4 looks very unstable to humans, significantly different from one with value 1. Figure~\ref{fig:boundary} shows examples with various visual instability values.

For this experiment, we restrict possible camera positions so that the deviations of blocks can be clearly perceived. 
We generate four datasets of stable blocks with visual instabilities of $1,2,3$, and $4$ respectively, each with 100 images. 

\smallvs\vspace{-5pt}
\paragraph{Results and Discussions}

As shown in Figure~\ref{fig:boundary}, the performance of neural networks are, in general, better than their performance in Experiment 1, probably because images here are easier as the camera positions are restricted. Also, their performance barely changes for groups with different visual instabilities. Even for the most deceptive group (visual instability 4), a LeNet has an accuracy of $93\%$. We also test AlexNet (both pretrained and not pretrained) on cases where blocks are unstable but visually stable, and the network, again, gives highly accurate results ($\geq 93\%$).

The performance of IPE and humans, on the other hand, changes drastically across groups. Corresponding to results in Experiment 1, both IPE and humans consistently predict that blocks with visual instability 4 will fall. Their accuracies are higher when visual instability is smaller, but still not close to those of neural networks. This confirms our observation of the asymmetry. More discussions follow in the final section.

\vspace{-2pt}
\section{Experiment 4: Knowledge Transfer}
\vspace{-2pt}

A possible explanation to humans' one-shot learning ability is based on the concept of transfer learning. In our fourth experiment, we evaluate the behaviors of computational models on tasks involving knowledge transfer. 

\smallvs\vspace{-5pt}
\paragraph{Experimental Setup}

\begin{figure}[t]
\centering
\includegraphics[width=0.24\linewidth]{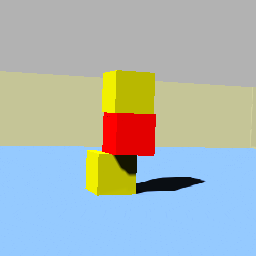}
\includegraphics[width=0.24\linewidth]{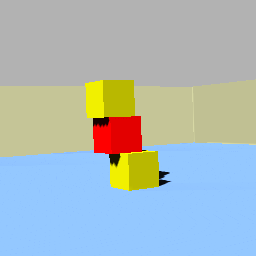}
\includegraphics[width=0.24\linewidth]{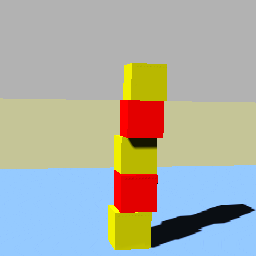}
\includegraphics[width=0.24\linewidth]{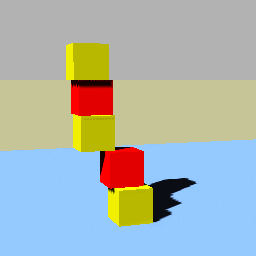}
\vspace{-10pt}
\caption{Images with three or five blocks}
\vspace{-15pt}
\label{fig:35blocks}
\end{figure}

For this experiment, we generate $200$ test images with three and five blocks, respectively. Examples are shown in Figure~\ref{fig:35blocks}. We modify the variance of block positions to ensure there are half stable and half unstable cases.

Our Bayesian vision system is extended to include the number of blocks as one parameter in sampling. 
Because the number of blocks directly determines the total mass, we also vary the magnitude of the perturbation force according to the inferred number of blocks to keep its effect consistent. 
For neural networks, we simply test the models previously trained on the $200,000$ images with four blocks.

\smallvs\vspace{-5pt}
\paragraph{Results and Discussions}

Table~\ref{table:transferRes} shows that while CNNs achieve $\sim90\%$ accuracies on four-block cases, their performance is much worse on cases where the number of blocks is smaller than that in training examples. Specifically, the predictions of models trained on $200$K images are at chance. For cases with more blocks, CNNs, especially pretrained AlexNet, can learn to generalize to some extent. However, their behaviors are different from human responses. 
In comparison, humans and the IPE model have relatively consistent performance, with slight decreases in accuracies as the number of blocks goes up and the task becomes more difficult.

These experiments demonstrate that the knowledge learned by neural networks cannot be transferred, at least in a straightforward way, to scenarios outside the training set. The IPE model and humans enjoy more flexibility in reasoning in the complex world and solving more general problems. 

\section{General Discussion}

\begin{table}[t]
\centering\small
\begin{tabular}{lccccc}
\toprule
\multirow{2}{*}{Model} & \multirow{2}{*}{Training} & \multicolumn{4}{c}{Test Set} \\
\cmidrule{3-6}
 & & 3 & 4 & 5 & Avg \\
\midrule
LeNet $(200$K$)$& 4 & 50.5 &88.5 &64.0 &67.7 \\
AlexNet $(200$K$)$ & 4 & 52.5 &89.5 & 65.5 & 69.2 \\
AlexNet $($P, $200$K$)$ & 4 & 51.0 &95.0 & 78.5 & 74.8 \\
\midrule
LeNet $(1,000)$ & 4 & 57.0 & 64.0 & 66.0 & 62.3 \\
AlexNet $(1,000)$ & 4 & 54.0 & 62.0 & 64.5 & 60.2 \\
AlexNet $($P, $1,000)$ & 4 & 55.0 & 71.0 & 72.0 & 66.0 \\
\midrule
IPE $(0.1, 10x)$ & N/A &72.0  & 64.0 & 56.0 & 64.0\\ 
Human & N/A & 76.5 & 68.5& 59.0 & 68.0 \\
\bottomrule
\end{tabular}
\vspace{-10pt}
\caption{Results on the task of transfer learning}
\label{table:transferRes}
\vspace{-15pt}
\end{table}

Following Facebook AI's reported results, we found that convolutional neural networks can be trained to achieve super-human accuracy levels on stability judgment tasks from raw images (Exps\onedot1 and 2).  CNNs also correlate reasonably well with human intuitions about how likely a stack of blocks is to fall, and once trained, they can respond to new images extremely quickly.  However, these features do not automatically make CNNs a good model of people's physical intuitions. They do not capture systematic judgment asymmetries that humans make, which simulation-based IPE models do capture (Exps\onedot1-3).  CNNs also have limited generalization ability across even small scene variations, such as changing the number of blocks. 
In contrast, IPE models naturally generalize and capture the ways that human judgment accuracy decreases with the number of blocks in a stack (Exp\onedot4).  

Taken together, these results point to something fundamental about human cognition that neural networks (or at least CNNs) are not currently capturing: the existence of a mental model of the world's causal processes.  Causal mental models can be simulated to predict what will happen in qualitatively novel situations, and they do not require vast and diverse training data to generalize broadly, but they are inherently subject to certain kinds of errors (\eg, propagation of uncertainty due to state and dynamics noise) just in virtue of operating by simulation.

Despite the success of CNNs in accounting for other high-level human perceptual capacities, such as rapid object classification \cite{yamins2014performance}, our results suggest that at least some perceptual judgments which people can make in a quick glance are not well explained by current feedforward neural networks.  We should not conclude however, that neural networks cannot help to explain how people make intuitive physical judgments. If people do indeed have a ``physics engine in the head'', somehow this simulator must be implemented in neural circuits.  Recurrent neural networks (RNNs) could provide one model for this \cite{fragkiadaki2015learning}. 
It is also possible that CNNs, if trained on more diverse scenes and physical judgments than those studied here and/or pretrained on large-scale image classification tasks (as in ~\citeNP{lerer2016learning}), could capture more of the qualitative inference behavior people show in our tasks.  Lastly, CNNs could be useful for visual intuitive physics by quickly estimating the relevant object properties in images needed to represent the world's state in a physics engine, which would then support more sophisticated reasoning and prediction by simulation~\cite{wu2015galileo}.  
Going forward we are eager to explore these and other productive lines of exchange between simulation-based generative models and memory-based neural network models. 

\noindent
{\bf Acknowledgement \;} We thank Tomer Ullman and Adam Lerer for helpful discussions. This work is in part supported by NSF RI 1212849 Reconstructive Recognition, CBMM (NSF STC award CCF-1231216), and MERL.

\renewcommand\bibliographytypesize{\small}
\bibliographystyle{apacite}

\setlength{\bibleftmargin}{.075in}
\setlength{\bibindent}{-\bibleftmargin}

\bibliography{blocks}

\end{document}